\documentclass[runningheads]{llncs}
\usepackage{graphicx}
\usepackage{tikz}
\usepackage{cite}
\usepackage{graphicx}
\usepackage{comment}
\usepackage{amsmath,amssymb} 
\usepackage{color}
\usepackage{float}
\usepackage[caption=false]{subfig}
\usepackage{epsfig}
\usepackage{xspace}
\usepackage{color}
\usepackage{graphics}
\usepackage{adjustbox}
\usepackage{amsmath}
\usepackage{algorithmicx,algorithm}
\usepackage{paralist} %
\usepackage{enumitem}
\usepackage{graphicx} %
\usepackage{epstopdf}%
\usepackage{booktabs} %

\usepackage[pagebackref=true,breaklinks=true,letterpaper=true,colorlinks,citecolor=blue,linkcolor=blue,bookmarks=false]{hyperref}

\long\def\ignorethis#1{}

\definecolor{Gray}{rgb}{0.35,0.35,0.35}
\definecolor{Blue}{rgb}{0,0.2,0.8}
\definecolor{Red}{rgb}{0.8,0.2,0}
\definecolor{Green}{rgb}{0.0,0.5,0.1}
\definecolor{Gray}{rgb}{0.4,0.4,0.4}

\def\red#1{\textcolor{Red}{#1}}
\def\blue#1{\textcolor{Blue}{#1}}

\newlength\paramargin
\newlength\figmargin
\newlength\secmargin

\usepackage{array}
\newcolumntype{L}[1]{>{\raggedright\let\newline\\\arraybackslash\hspace{0pt}}m{#1}}
\newcolumntype{C}[1]{>{\centering\let\newline\\\arraybackslash\hspace{0pt}}m{#1}}
\newcolumntype{R}[1]{>{\raggedleft\let\newline\\\arraybackslash\hspace{0pt}}m{#1}}

\def\eg{e.g.,~}

\graphicspath{{figure/eps/}}

\usepackage{tabulary,multirow,overpic,xcolor}

\newcommand{\ve}[1]{\mathbf{#1}} %

\newcolumntype{x}[1]{>{\centering\arraybackslash}p{#1pt}}
\newcommand{\tablestyle}[2]{\setlength{\tabcolsep}{#1}\renewcommand{\arraystretch}{#2}\centering\footnotesize}
\newlength\savewidth\newcommand\shline{\noalign{\global\savewidth\arrayrulewidth
		\global\arrayrulewidth 1pt}\hline\noalign{\global\arrayrulewidth\savewidth}}

\makeatletter\renewcommand\paragraph{\@startsection{paragraph}{4}{\z@}
	{.5em \@plus1ex \@minus.2ex}{-.5em}{\normalfont\normalsize\bfseries}}\makeatother
	
\newcommand{\bd}[1]{\textbf{#1}}

\definecolor{orange}{rgb}{1,0.5,0}

\begin{document}
\pagestyle{headings}
\mainmatter
\def\ECCVSubNumber{241}  %

\title{Representative Graph Neural Network} %

\author{
    Changqian Yu\inst{1,2}
    \and
    Yifan Liu\inst{2}
    \and
    Changxin Gao\inst{1}
    \and
    Chunhua Shen\inst{2}
    \and \\
    {Nong Sang}\inst{1}\thanks{Corresponding author. 
    Part of the work was done when C. Yu was visiting The University of Adelaide.}
}

\institute{Key Laboratory of Image Processing and Intelligent Control, \\
School of Artificial Intelligence and Automation, \\
Huazhong University of Science \& Technology, China \and
The University of Adelaide, Australia \\
\email{\{changqian\_yu,cgao,nsang\}@hust.edu.cn}}

\authorrunning{C. Yu et al.}
\maketitle

\begin{abstract}
Non-local operation is widely explored to model the long-range dependencies.
However, the redundant computation in this operation leads to a prohibitive complexity.
In this paper, we present a Representative Graph (RepGraph) layer to dynamically sample a few representative features, which dramatically reduces redundancy.
Instead of propagating the messages from all positions, our RepGraph layer computes the response of one node merely with a few representative nodes.
The locations of representative nodes come from a learned spatial offset matrix.
The RepGraph layer is flexible to integrate into many visual architectures and combine with other operations.
With the application of semantic segmentation, without any bells and whistles, our RepGraph network can compete or perform favourably against the state-of-the-art methods on three challenging benchmarks: ADE20K, Cityscapes, and PASCAL-Context datasets.
In the task of object detection, our RepGraph layer can also improve the performance on the COCO dataset compared to the non-local operation.
Code is available at \url{https://git.io/RepGraph}.
\keywords{Representative Graph; Dynamic Sampling; Semantic Segmentation; Deep Learning}
\end{abstract}

\section{Introduction}
\label{sec:intro}
Modelling long-range dependencies is of vital importance for visual understanding, \eg semantic segmentation~\cite{Yuan-Arxiv-OCNet-2018, Fu-CVPR-DANet-2019, Huang-ICCV-CCNet-2019} and object detection/segmentation~\cite{Wang-CVPR-Nonlocal-2018, Hu-CVPR-RelationNet-2018, Cao-ARXIV-GCNet-2019}.
The previous dominant paradigm is dependent on the deep stacks of local operators, \eg convolution operators, which are yet limited by inefficient computation, hard optimization and insufficient effective receptive field~\cite{Luo-NIPS-ERF-2016}.

\begin{figure}[t]
\footnotesize
\centering
\renewcommand{\tabcolsep}{2pt} %
\renewcommand{\arraystretch}{1} %
\begin{center}
\begin{tabular}{cc}
\includegraphics[width=0.35\textwidth]{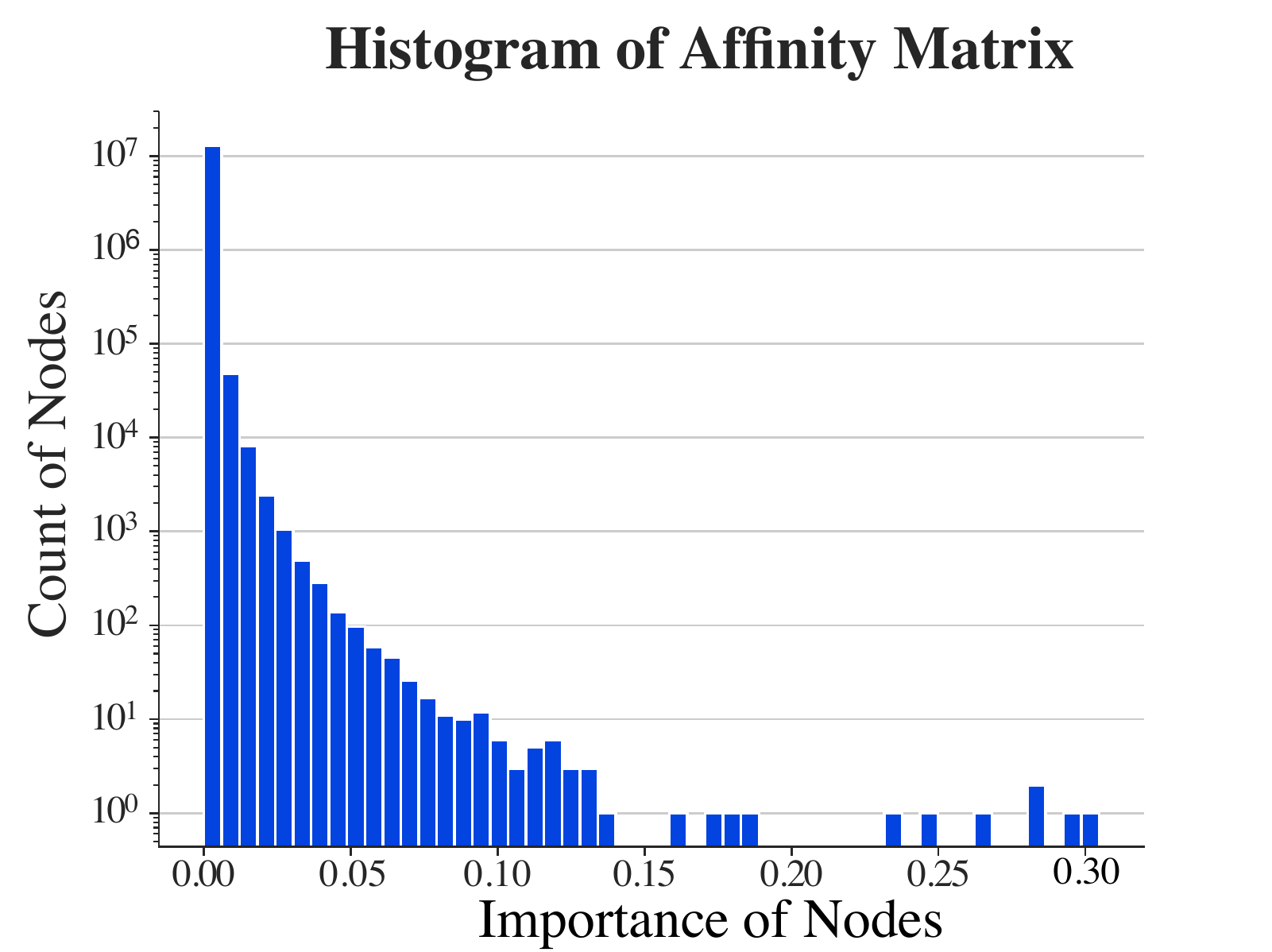}  & 
\includegraphics[width=0.35\textwidth]{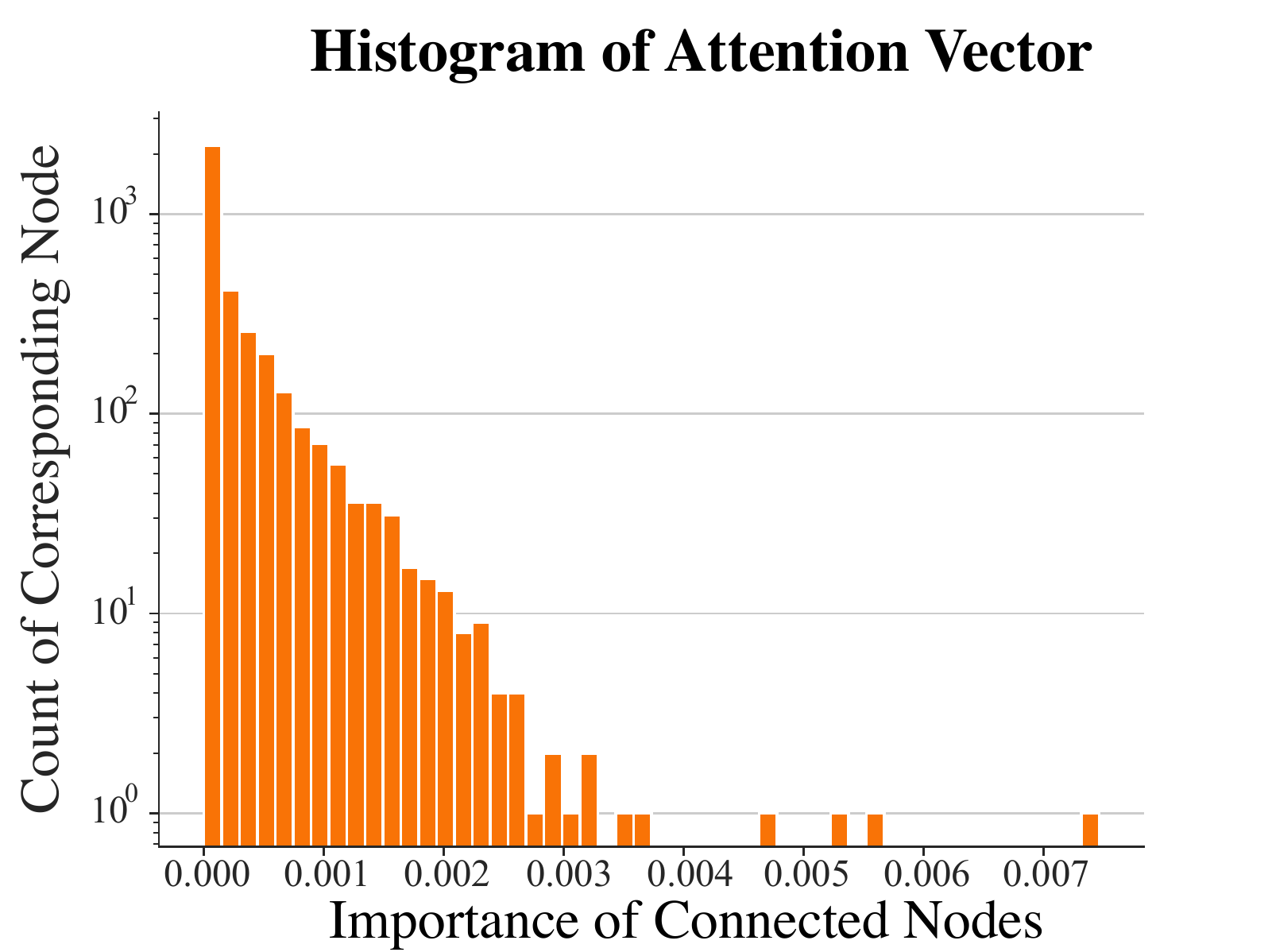} \\
\multicolumn{2}{c}{(a) Statistical analysis of the affinity matrix in the non-local operation} \\
\multicolumn{2}{c}{\includegraphics[width=0.7\textwidth]{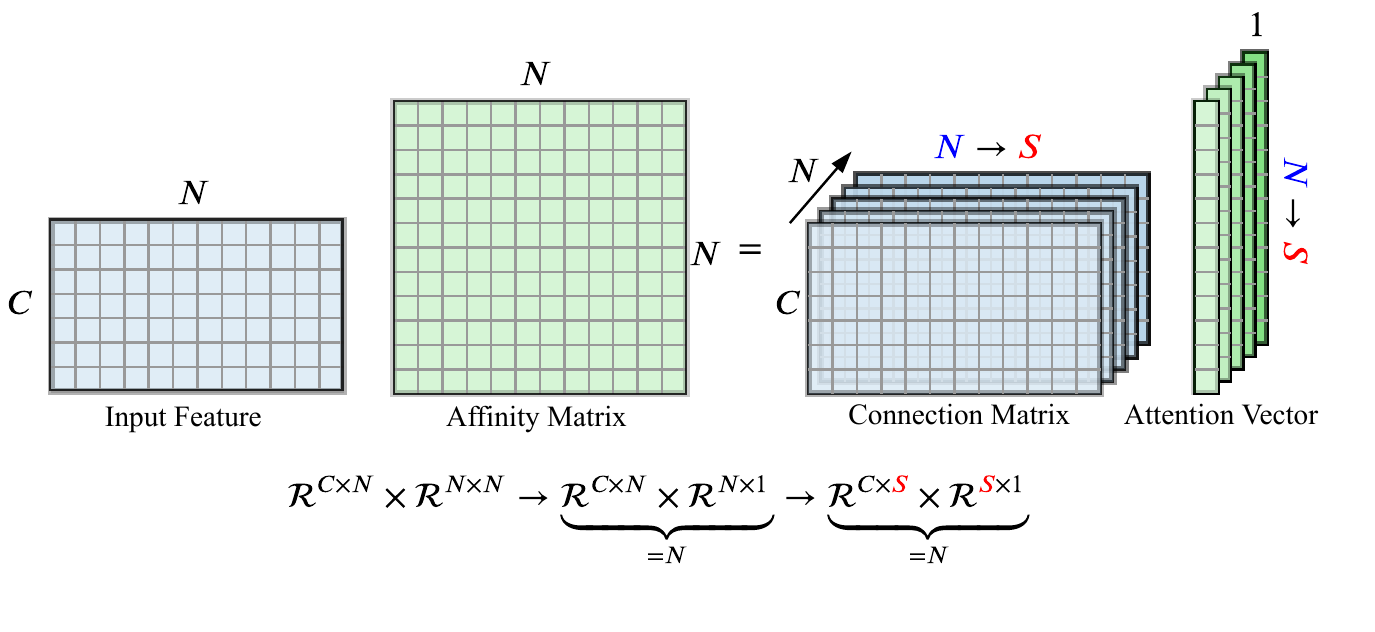}}\\
\multicolumn{2}{c}{(b) Computation diagram of the non-local operation} \\
\end{tabular}
\end{center}
\caption{\textbf{Illustration of the affinity matrix.} 
(a) The statistical analysis of the affinity matrix. In Non-local, the affinity matrix models the importance of all other positions to each position. The histogram of the weights of all the positions is on the left, while that of one position is on the right. These statistical results indicate some \emph{representative} positions exhibit principal importance.
(b) The computation diagram of Non-local. The computation in the left can be decoupled into $N$ groups of connection matrix and attention vector. The connection matrix determines which positions require connecting to the current position, while the attention vector assigns the weight to the connection edge. Here, the $\blue{N}$ above the connection matrix represents $\blue{N}$ nodes are connected to the current position. Therefore, for each position, we can select $\red{S}$ representative nodes to connect and compute a corresponding attention vector to reduce the complexity}
\label{fig:fig1}
\end{figure}

To solve this issue, a family of non-local methods~\cite{Wang-CVPR-Nonlocal-2018, Hu-CVPR-RelationNet-2018, Yuan-Arxiv-OCNet-2018, Zhang-CVPR-ACF-2019, Chen-CVPR-GloRe-2019, Yu-CVPR-CPN-2020} is proposed to capture the long-range relations in feature representation. 
The non-local operation computes the response at a position as a weighted sum of the features at all positions, whose weights are assigned by a dense affinity matrix.
This affinity matrix models the importance of all other positions to each position.
Intuitively, there is some redundancy in this affinity matrix, leading to a prohibitive computation complexity.
For each position, some other positions may contribute little to its response.
This assumption drives us to study further the distributions of importance in the affinity matrix.
We perform the statistical analysis on this affinity matrix to help to understand in Figure~\ref{fig:fig1} (a).
It is surprising to see that the distribution is extremely imbalanced, which implies that \textit{some representative positions contribute principal impact, while the majority of positions have little contribution.} 
Therefore, if the affinity matrix can only compute the response with only a few representative positions, the computation redundancy can be dramatically reduced. 

Based on previous observations, we rethink Non-local methods from the graphical model perspective and propose an efficient, flexible, and generic operation to capture the long-range dependencies, termed \textbf{Rep}resentative \textbf{Graph} (RepGraph).
Considering each feature position as a node, Non-local constructs a complete graph (fully-connected graph) and assigns a weight to each connection edge via the affinity map.
Thus, for each feature node, the computation of non-local graph can be decoupled into two parts: (\romannumeral1) a connection matrix; (\romannumeral2) a corresponding attention vector, as illustrated in Figure~\ref{fig:fig1}~(b).
The connection matrix determines which nodes make contributions to each node, while the attention vector assigns the weight for each connection edge.

To reduce the redundant computation, for each node, the \textit{RepGraph} layer dynamically selects a few representative nodes as the neighbourhoods instead of all nodes. The corresponding weight is assigned to the edge to propagate the long-range dependencies.
Specifically, the \textit{RepGraph} layer first regresses an offset matrix conditioned on the current node.
With the offset matrix, this layer samples the representative nodes on the graph, and then compute an affinity map as the weight to aggregate these representative features.

Meanwhile, the \textit{RepGraph} layer is easy to combine with other mechanisms.
Motivated by the pyramid methods~\cite{Zhao-CVPR-PSPNet-2017, Chen-Arxiv-Deeplabv2-2016}, we can spatially group some feature elements as a graph node instead of only considering one element, named \textit{Grid Representative Graph (Grid RepGraph)}.
Besides, inspired by the channel group mechanism~\cite{Xie-CVPR-ResNeXt-2017, Howard-Arxiv-MobileNet-2017, Sandler-Arxiv-MobileNetv2-2018, Zhang-CVPR-Shufflenet-2018, Yue-NIPS-CGNL-2018}, we can also divide the features into several groups, and conduct computation in each corresponding group, termed \textit{Group RepGraph}.

There are several merits of our \textit{RepGraph} layer: 
(i) The \textit{RepGraph} layer can dramatically reduce the redundant computation of Non-local. 
Specifically, the RepGraph layer is around $17$ times smaller in computation complexity and over $5$ times faster than Non-local with a $256 \times 128$ input.
(ii) The \textit{RepGraph} layer learns a compact and representative feature representation, which is more efficient and effective than non-local operation;
(iii) Our \textit{RepGraph} layer can be flexibly integrated into other neural networks or operations.

We showcase the effectiveness of \textit{RepGraph} operations in the application of semantic segmentation. 
Extensive evaluations demonstrate that the \textit{RepGraph} operations can compete or perform favorably against the \textit{state-of-the-art} semantic segmentation approaches.
To demonstrate the generality of \textit{RepGraph} operations, we further conduct the experiments of object detection/segmentation tasks on the COCO dataset~\cite{Lin-COCO-2014}.
In the comparison of non-local blocks, our \textit{RepGraph} operations can increase the accuracy further.

\section{Related Work}
\label{sec:related-work}
\paragraph{Non-local methods and compact representation.}
Motivated by the non-local means~\cite{Buades-CVPR-NLM-2005}, \cite{Wang-CVPR-Nonlocal-2018} proposes the non-local neural network to model the long-range dependences in the application of video classification~\cite{Wang-CVPR-Nonlocal-2018, Chen-NIPS-A2Net-2018}, object detection and segmentation~\cite{Wang-CVPR-Nonlocal-2018, Hu-CVPR-RelationNet-2018, Zhang-CVPR-ACF-2019}.
The relation network~\cite{Hu-CVPR-RelationNet-2018} embeds the geometry feature with the self-attention manner~\cite{Ashish-NIPS-Transfomer-2017} to model the relationship between object proposals.
OCNet~\cite{Yuan-Arxiv-OCNet-2018} extends the non-local block with pyramid methods~\cite{Chen-Arxiv-Deeplabv3-2017, Zhao-CVPR-PSPNet-2017}.
CFNet~\cite{Zhang-CVPR-ACF-2019} explores the co-occurrent context to help the scene understanding, while DANet~\cite{Fu-CVPR-DANet-2019} applies the self-attention on both the spatial and channel dimension.
Meanwhile, CPNet~\cite{Yu-CVPR-CPN-2020} explores the supervised self-attention matrix to capture the intra-context and inter-context.

The redundant computation in these nonlocal methods leads to a prohibitive computational complexity, which hinders its application, especially in some dense prediction tasks.
Therefore, there are mainly two ways to explore the compact representation of non-local operations:
(\romannumeral 1) Matrix factorization.
$A^2$-Net~\cite{Chen-NIPS-A2Net-2018} computes the affinity matrix between channels.
LatentGNN~\cite{Zhang-ICML-LatentGNN-2019} embeds the features into a latent space to get a low-complexity matrix.
CGNL~\cite{Yue-NIPS-CGNL-2018} groups the channels and utilizes the Taylor expansion reduce computation.
CCNet~\cite{Huang-ICCV-CCNet-2019} introduces a recurrent criss-cross attention.
(\romannumeral 2) Input restricting.
Non-local has a quadratic complexity with the size of the input feature.
Therefore, restricting the input size is a straightforward approach to reduce complexity.
\cite{Zhu-ICCV-ANL-2019} utilizes a pyramid pooling manner to compress the input of the key and value branch.
ISA~\cite{Huang-ARXIV-ISA-2019} adopts the interlacing mechanism to spatially group the input.

In this paper, we explore a compact and general representation to model the dependencies. 
Non-local can be a special case of our work, as illustrated in Figure~\ref{fig:fig1} (b).
In contrast to previous compact methods, our work dynamically sample the representative nodes to efficiently reduces the spatial redundancy.

\paragraph{Graph neural network.}
Our work is also related to the graphical neural network~\cite{Lafferty-ICML-CRF-2001, Krahenbuhl-NIPS-CRF-2011, Zheng-ICCV-CRFasRNN-2015}.
Non-local~\cite{Ashish-NIPS-Transfomer-2017, Wang-CVPR-Nonlocal-2018} can be viewed as a densely-connected graph, which models the relationships between any two nodes.
Meanwhile, GAT~\cite{Velivkovi-ICLR-GAT-2018} introduces a graph attentional layer, which performs self-attention on each node.
In contrast to both dense graphical models, our work constructs a sparse graph, on which each node is simply connected to a few representative nodes.

\paragraph{Deformable convolution.}
Our work needs to learn an offset matrix to locate some representative nodes, which is related to the deformable convolution~\cite{Dai-ICCV-DCN-2017}.
The learned offset matrix in DCN is applied on the regular grid positions of convolution kernels, and the number of the sampled positions requires matching with the kernel size.
However, our work applies the learned offset to each node position directly.
The number of sampled nodes can be unlimited theoretically.

\section{Representative Graph Neural Networks}
\label{sec:method}
In this section, we first revisit Non-local from the graphical model perspective in Section~\ref{sec:method:non-local}.
Next, we introduce the motivation, formulation, and several instantiations of the representative graph layer in Section~\ref{sec:method:rep-graph}.
Finally, the extended instantiations of the representative graph layer are illustrated in Section~\ref{sec:method:extended}.

\begin{figure}[t]
\footnotesize
\centering
\renewcommand{\tabcolsep}{2em} %
\renewcommand{\arraystretch}{1} %
\begin{center}
\begin{tabular}{cc}
\includegraphics[width=0.25\textwidth]{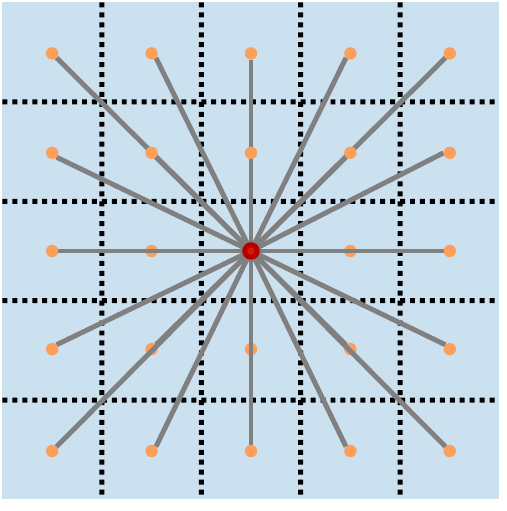}  & 
\includegraphics[width=0.25\textwidth]{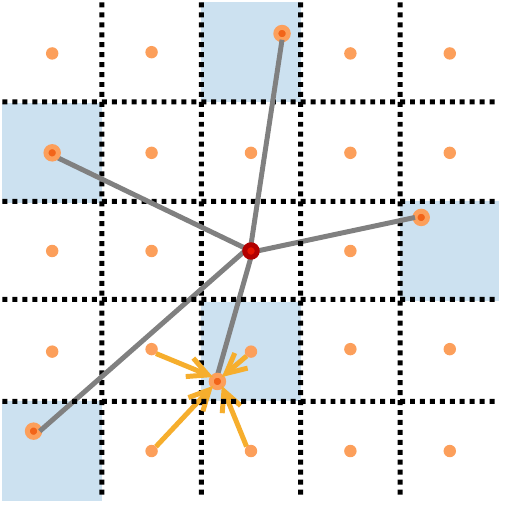} \\
(a) Non-local graph & (b) Representative graph
\end{tabular}
\end{center}
\caption{\textbf{Comparison of non-local graph and representative graph.}
(a) The non-local graph is a fully-connected graph, leading to prohibitive complexity.
(b) The representative graph only computes the relationships with some representative nodes, \eg five in the figure, for each output node. The sampled nodes with fractional positions are interpolated  with the nodes at four integral neighbourhood positions
}
\label{fig:graph-view}
\end{figure}

\subsection{Revisiting Non-local Graph Neural Network}
\label{sec:method:non-local}
Following the formulation of the non-local operator in \cite{Wang-CVPR-Nonlocal-2018}, we describe a fully-connected graphical neural network.

For a 2D input feature with the size of $C \times H \times W$, where $C$, $H$, and $W$ denote the channel dimension, height, and width respectively, it can be interpreted as a set of features, $\ve{X} = [\ve{x}_1,\ve{x}_2, \cdots, \ve{x}_N]^\intercal$, $\ve{x}_i\in\mathbb{R}^C$, where $N$ is the number of nodes (\eg $N=H \times W$), and $C$ is the node feature dimension.

With the input features, we can construct a fully-connected graph $\mathcal{G}=(\mathcal{V},\mathcal{E})$ with $\mathcal{V}$ as the nodes, and $\mathcal{E}$ as the edges, as illustrated in Figure~\ref{fig:graph-view} (a).
The graphical model assigns each feature element $\ve{x}_i$ as a node $v_i \in \mathcal{V}$, while the edge $(v_i, v_j) \in \mathcal{E}$ encodes the relationship between node $v_i$ and node $v_j$.
Three linear transformation, parameterized by three weight matrices $W_\phi \in \mathbb{R}^{C' \times C}$, $W_\theta \in \mathbb{R}^{C' \times C}$, and $W_g \in \mathbb{R}^{C' \times C}$ respectively, are applied on each node.
Therefore, the formulation of the non-local graph network can be interpreted as:
\begin{equation}\label{eq:nonlocal}
\begin{aligned}
\ve{\tilde{x}}_i = \frac{1}{\mathcal{C(\ve{x})}} \sum_{\forall j}f(\ve{x}_i, \ve{x}_j)g(\ve{x}_j) 
= \frac{1}{\mathcal{C(\ve{x})}} \sum_{\forall j} \delta(\theta(\ve{x}_i)\phi(\ve{x}_j)^\intercal) g(\ve{x}_j),
\end{aligned}
\end{equation}
where $j$ enumerates all possible positions, $\delta$ is the softmax function, and $\mathcal{C(\ve{x})}$ is a normalization factor.

We can rewrite the formulation in Equation~\ref{eq:nonlocal} in a matrix form:
\begin{equation}\label{eq:nonlocal-matrix}
\begin{aligned}
\ve{\tilde{X}} = \delta(\ve{X_\theta}\ve{X_\phi}^\intercal)\ve{X_g} 
=\ve{A}(\ve{X})\ve{X_g},
\end{aligned}
\end{equation}
where $\ve{A}(\ve{X}) \in \mathbb{R}^{N \times N}$ indicates the affinity matrix, $X_\theta \in \mathbb{R}^{N \times C'}$, $X_\phi \in \mathbb{R}^{N \times C'}$, and $X_g \in \mathbb{R}^{N \times C'}$.
The matrix multiplication results in a prohibitive computation complexity: $\mathcal{O}(C' \times N^2)$.
In some visual understanding tasks, \eg semantic segmentation and object detection, the input usually has a large resolution, which is unfeasible to compute the dense affinity matrix.
It is thus desirable to explore a more compact and efficient operation to model the long-range dependencies.

\begin{figure}[t]
\footnotesize
\centering
\renewcommand{\tabcolsep}{2pt} %
\renewcommand{\arraystretch}{1} %
\begin{center}
\begin{tabular}{c}
\includegraphics[width=0.99\textwidth]{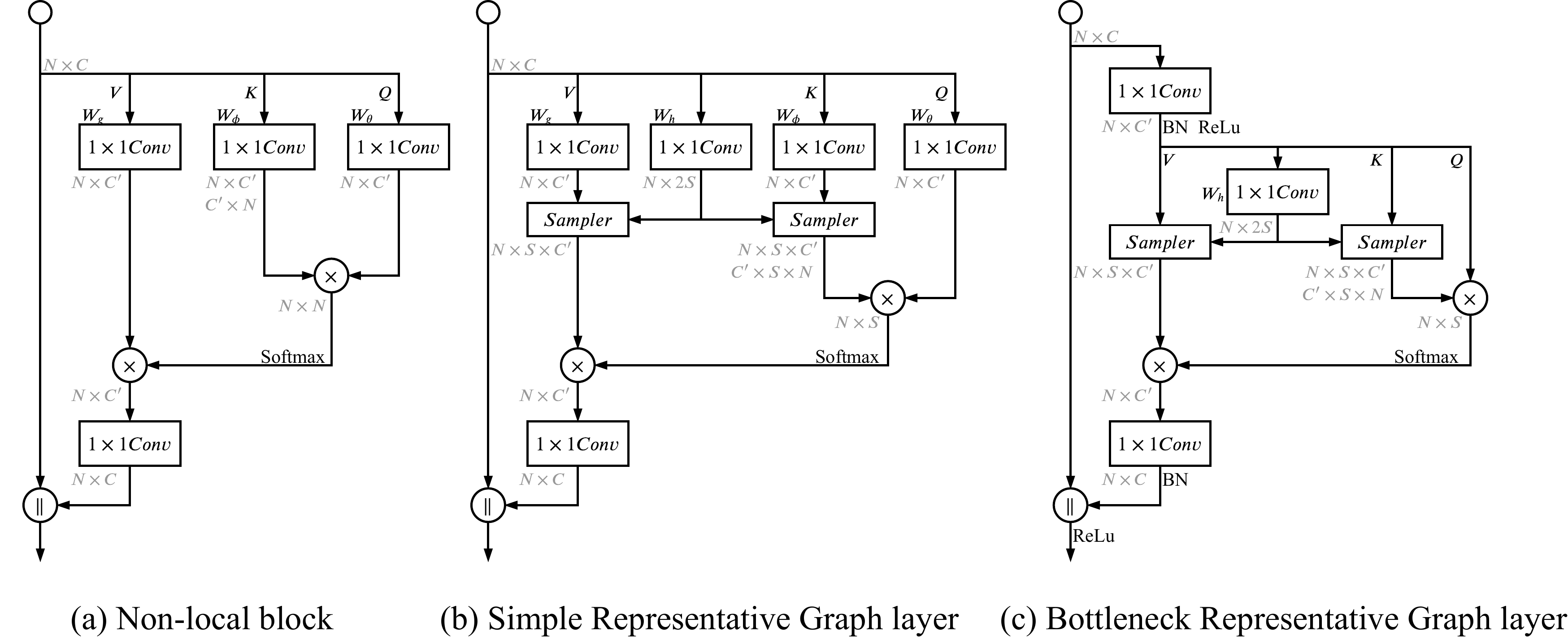} 
\end{tabular}
\end{center}
\caption{\textbf{Instantiations of Representative Graph layer.}
(a) is the structure of the non-local block;
(b) is the structure of our Simple RepGraph layer, which utilizes learned offset matrix to sample the representative features of the value and key branches;
(c) is the structure of our Bottleneck RepGraph layer, following the bottleneck design.
Here, $\parallel$ means the aggregation operation, \eg \emph{summation} or \emph{concatenation}}
\label{fig:structure}
\end{figure}

\subsection{Representative Graph Layer}
\label{sec:method:rep-graph}
\paragraph{Motivation.}
As showed in Figure~\ref{fig:fig1} (b), we can reconstruct the matrix multiplication as a new form:
	\begin{equation}
	\mathcal{R}^{{N} \times N} \times  \mathcal{R}^{N \times C'} \rightarrow \underbrace{\mathcal{{\color{orange}{R}}}^{{1} \times N} \times  \mathcal{{\color{blue}{R}}}^{N \times C'}}_{=N},
	\end{equation}
where $\mathcal{{\color{blue}{R}}}^{N \times C'}$, as the connection matrix, determines which nodes are connected to current node, while $\mathcal{{\color{orange}{R}}}^{1 \times N}$ as the attention vector assigns the weight to corresponding edge.

Based on our observation discussed in Section~\ref{sec:intro}, we can select a few representative nodes (\eg $S$) for each node instead of propagating the messages from all nodes.
Therefore, the number of connected nodes for each node can be reduced from $N$ to $S$ (usually $S \ll N$), which dramatically reduces the prohibitive computation complexity.
The new pipeline can be transformed as:
	\begin{equation}
	\begin{aligned}
	\underbrace{\mathcal{{\color{orange}{R}}}^{{1} \times N} \times  \mathcal{{\color{blue}{R}}}^{N \times C'}}_{=N} & \rightarrow
	\underbrace{\mathcal{{\color{orange}{R}}}^{{1} \times {\color{red}{S}}} \times  \mathcal{\color{blue}{R}}^{{\color{red}{S}} \times C'}}_{=N}, \\
	\end{aligned}
	\end{equation}
where $S$ is the number of the representative nodes.
This reconstruction reduces the computation cost from $\mathcal{O}(C' \times N^2)$ to $\mathcal{O}(C' \times N \times S)$, usually $S \ll N$ (\eg for a $65 \times 65$ input feature, $N = 65 \times 65 = 4225$, while $S = 9$ in our experiments).

\begin{figure}[t]
\footnotesize
\centering
\begin{center}
\includegraphics[width=0.5\textwidth]{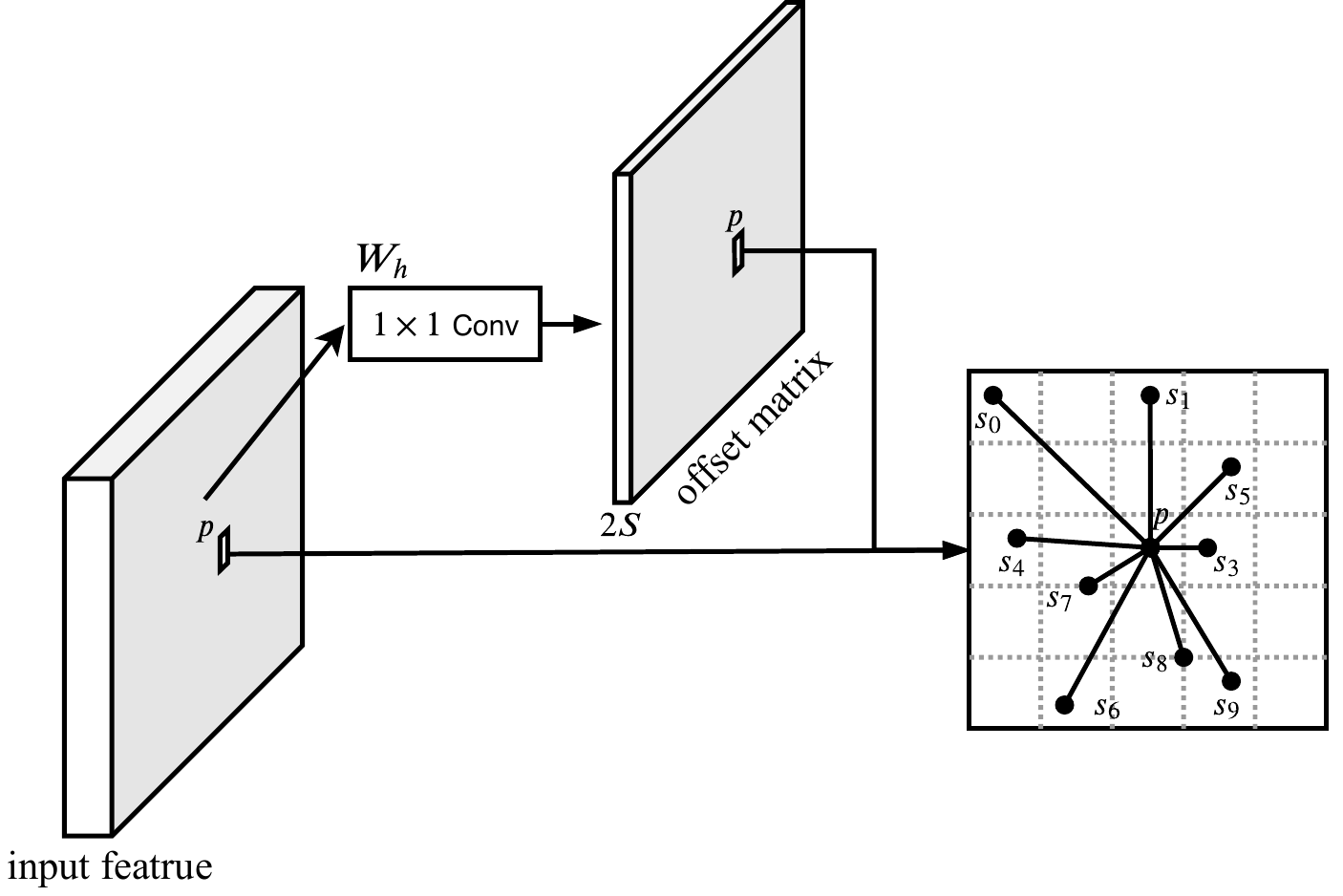}
\end{center}
\caption{\textbf{Illustration of representative nodes sampler.} For each position $p$, the layer adopts one $1 \times 1$ convolution operation to regress an offset matrix to sample $S$ representative nodes. The offset matrix has $2S$ channels, the values of which are typically fractional}
\label{fig:offset}
\end{figure}

\paragraph{Formulation.}
Based on the observations, we can dynamically sample some nodes to construct the Representative Graph, as illustrated Figure~\ref{fig:graph-view} (b).

For each position $i$ of the node feature, we sample a set of representative node features:
\begin{equation}
\label{eq:sample-func}
	\mathcal{F}(\ve{x}(i)) = \left\{ \ve{s}(n)|n=1,2,\cdots, S \right\},
\end{equation}
where $s(n) \in \mathcal{R}^{C'}$ is the sampled representative node feature.

Given the sampling function $\mathcal{F}(\ve{x}_i)$, Equation~\ref{eq:nonlocal} can be reformulated as:
\begin{equation}
\label{eq:rep-graph}
\begin{aligned}
\ve{\tilde{x}}(i) &= \frac{1}{\mathcal{C(\ve{x})}} \sum_{\forall n} \delta(\ve{x}_{\theta}(i)\ve{s}_{\phi}(n)^\intercal) \ve{s}_g(n), \\
\end{aligned}
\end{equation}
where $n$ only enumerates the sampled positions, $\delta$ is the softmax function, $\ve{x}_{\theta}(i) = W_{\theta}\ve{x}(i), \ve{s}_{\phi}(i) = W_{\phi}\ve{s}(i), \ve{s}_{g}(i) = W_{g}\ve{s}(i)$.

\paragraph{Representative nodes sampler.}
Motivated by \cite{Dai-ICCV-DCN-2017}, we can instantiate Equation~\ref{eq:sample-func} via offset regression.
Conditioned on the node features, we can learn an offset matrix to dynamically select nodes.
Therefore, for each position $p$, Equation~\ref{eq:sample-func} can be reformulated as:
\begin{equation}
\begin{aligned}
	\mathcal{F}(\ve{x}(p)) &= \left\{ \ve{x}(p + \Delta p_n)|n=1,2,\cdots, S \right\}, 
\end{aligned}
\end{equation}
where $\Delta p_n$ is the regressed offset.

Due to the regression manner, the offset $\Delta p_n$ is commonly fractional.
Thus, we utilize bilinear interpolation~\cite{Jaderbera-NIPS-STN-2015} to compute the correct values of the fractional position with the node feature at four integral neighbourhood positions:
\begin{equation}
	\ve{x}(p_s) = \sum_{\forall t}G(t, p_s) \ve{x}(t),
\end{equation} 
where $p_s = p + \Delta p_n$, $t$ is four neighbourhood integral positions, and $G$ is bilinear interpolation kernel.

As illustrated in Figure~\ref{fig:offset}, we adopt a $1 \times 1$ convolutional layer to regress the offset matrix for each node feature, which has $2S$ channel dimensions.
After bilinear interpolation, the RepGraph layer can sample $S$ representative nodes.

\paragraph{Instantiations.}
We can instantiate Equation~\ref{eq:rep-graph} with a residual structure~\cite{He-CVPR-ResNet-2016}, as illustrated in Figure~\ref{fig:structure} (b).
We define the RepGraph layer as:
\begin{equation}
	\ve{y}_i = W_y \ve{\tilde{x}}_i + \ve{x}_i ,
\end{equation}
where $\ve{\tilde{x}}_i$ is given in Equation~\ref{eq:rep-graph}, and $\ve{x}_i$ is the original input feature.
This residual structure enables the RepGraph layer can be inserted to any pre-trained model.
We note that when applied into the pre-trained model, $W_y$ should be initialized to zero in avoid of changing the initial behavior of the pre-trained model.

As shown in Figure~\ref{fig:structure} (b), the RepGraph layer adopts a $1 \times 1$ convolution layer to regress the offset matrix, and sample the representative nodes of the key and value branch.
The features of the query branch conduct the matrix multiplication with the sampled representative node features of the key branch to obtain the attention matrix.
Then the attention matrix assigns corresponding weights and aggregates the representative node features of the value branch.

Meanwhile, motivated by the bottleneck design~\cite{He-CVPR-ResNet-2016, Xie-CVPR-ResNeXt-2017}, we re-design the RepGraph layer as a bottleneck structure, as illustrated in Figure~\ref{fig:structure} (c).
We note that when applied to the pre-trained architectures, the weights and biases of last convolution and batch normalization should be initialized to zero, and the ReLU function should be removed.

\begin{figure}[t]
\footnotesize
\centering
\renewcommand{\tabcolsep}{2pt} %
\renewcommand{\arraystretch}{1} %
\begin{center}
\begin{tabular}{cc}
\includegraphics[width=0.5\linewidth]{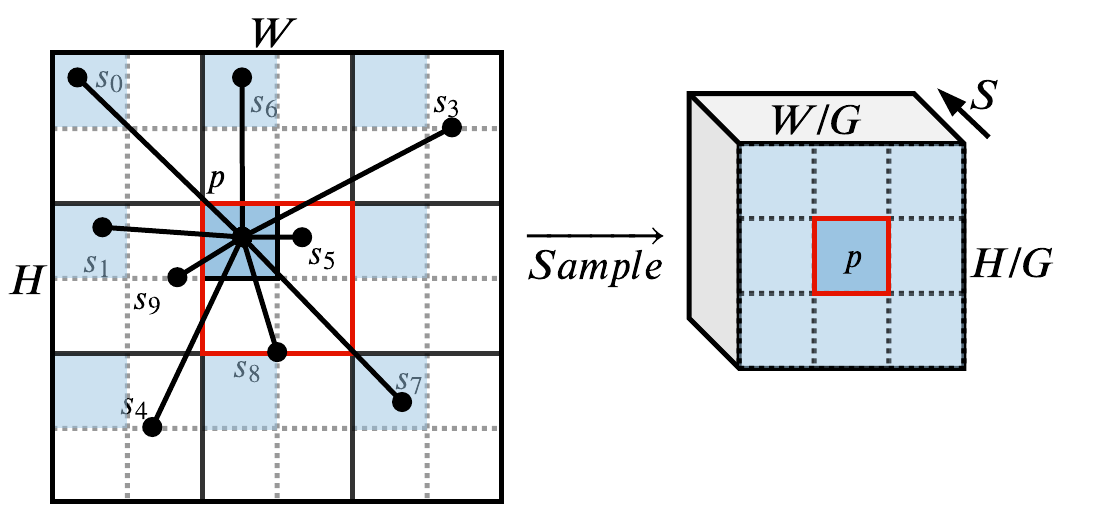} &
\includegraphics[width=0.45\linewidth]{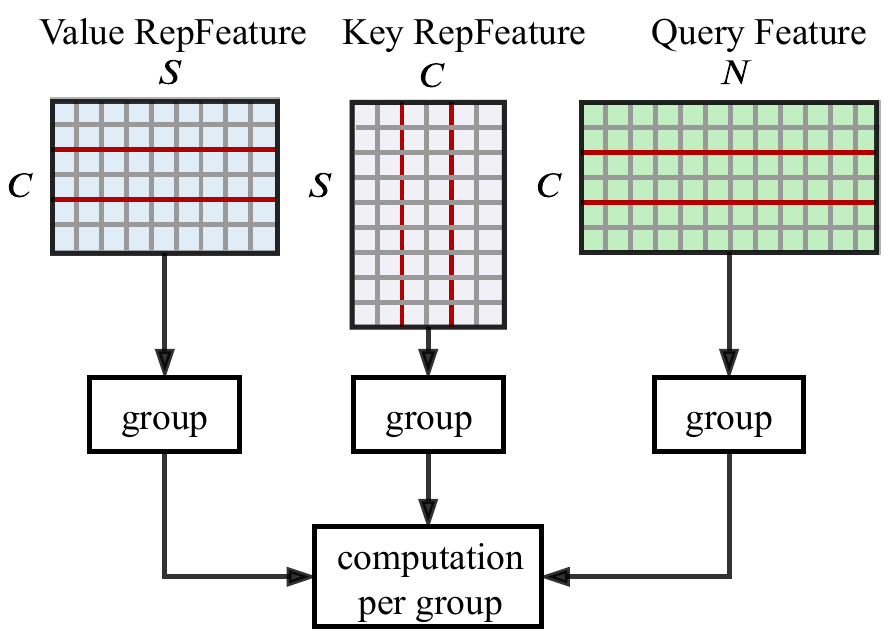} \\
(a) Grid RepGraph Layer &
(b) Group RepGraph Layer \\ 
\end{tabular}
\end{center}
\caption{\textbf{Illustration of extended instantiations of the RepGraph layer.} 
(a) is Grid RepGraph layer, which spatially grids the input features into several groups, \eg the features in the {\color{red}{red}} box. Each group has an anchor coordinate, showed in {\color{blue}{blue}} cube. The learned offset matrix is applied on the anchor coordinate to sample $S$ representative node features. $G$ indicates how many elements to group along a dimension, \eg $G=2$ in the top figure.
(b) is Group RepGraph layer, which divides the feature of query branch, sampled representative feature of value branch and key branch into several channel groups respectively. Then, the same computation shown in Equation~\ref{eq:rep-graph} is conducted in each corresponding group}
\label{fig:offset}
\end{figure}

\subsection{Extended Instantiations}
\label{sec:method:extended}
Motivated by pyramid methods~\cite{Zhao-CVPR-PSPNet-2017, Chen-Arxiv-Deeplabv2-2016}, and channel group mechanism~\cite{Xie-CVPR-ResNeXt-2017, Howard-Arxiv-MobileNet-2017, Yue-NIPS-CGNL-2018}, it is easy to instantiate the RepGraph layer with diverse structures.

\paragraph{Grid RepGraph layer.}
Instead of considering one input feature element as a node, we can spatially group varying quantity of elements as a node.
First, we spatially grid input feature into several groups.
The left-top element in each group is the anchor position. 
Then, we utilize the average pooling to group the input features to regress the offset matrix spatially.
The learned offset coordinates are applied to the anchor positions to sample some representative nodes for each group.
Finally, we conduct the matrix multiplication on the grid features. 

\paragraph{Group RepGraph layer.}
Channel group mechanism is widely used in light-weight recognition architectures, which can reduce the computation and increase the capacity of the model~\cite{Xie-CVPR-ResNeXt-2017, Howard-Arxiv-MobileNet-2017, Sandler-Arxiv-MobileNetv2-2018, Zhang-CVPR-Shufflenet-2018, Yue-NIPS-CGNL-2018}.
It is easy to be applied on the RepGraph layer.
With the input features $\ve{x}_{\theta}(i), \mathcal{F}(\ve{x}_{\phi}(j)), \mathcal{F}(\ve{x}_g(j))$, we can divide all $C'$ channels into $G$ groups, each of which has $\tilde{C} = C' / G$ channels.
Then, the RepGraph computation of Equation~\ref{eq:rep-graph} can be performed in each group independently.
Finally, we concatenate the output features of all the groups along the dimension of channels as the final output features.

\section{Experiments on Semantic Segmentation}
\label{sec:exp}
We perform comprehensive ablative evaluation on challenging ADE20K~\cite{Zhou-ADE-2016} dataset.
We also report performance on Cityscapes~\cite{Cityscapes} and PASCAL-Context~\cite{PASCAL-Context} to investigate the effectiveness of our work.

\paragraph{Datasets.} The ADE20K dataset contains 20K training images and 2K validation images. 
It is a challenging scene understanding benchmark due to the complex scene and up to 150 category labels. 

Cityscapes is a large urban street scene parsing benchmark, which contains 2,975, 500, 1,525 fine-annotation images for training, validation, and testing, respectively. 
Besides, there are additional 20,000 coarse-annotation images for training.
In our experiments, we only use the fine-annotation set. 
The images of this benchmark are all in $2048 \times 1024$ resolution with $19$ semantic categories.

PASCAL-Context~\cite{PASCAL-Context} augments 10,103 images from PASCAL VOC 2010 dataset~\cite{Pascal-VOC-2012} for scene understanding, which considers both the stuff and thing categories.
This dataset can be divided into 4,998 images for training and 5,105 images for testing. The most common 59 categories are used for evaluation.

\paragraph{Training.} We utilize the SGD algorithm with $0.9$ momentum to fine-tune the RepGraph network.
For the ADE20K and PASCAL-Context datasets, we train our model starting with the initial learning rate of $0.02$, the weight decay of $0.0001$, and the batch size of $16$.
For the Cityscapes dataset, the initial rate is $0.01$ with the weight decay of $0.0005$, while the batch size is $8$.
We note that we adopt the ``poly'' learning rate strategy~\cite{Chen-Arxiv-Deeplabv2-2016}, in which the initial learning rate is multiplied by $(1 - \frac{iter}{iter_{max}})^{0.9}$. 
Besides, the synchronized batch normalization~\cite{Zhang-CVPR-EncNet-2018, Peng-CVPR-MegDet-2018, Ioffe-ICML-BN-2015} is applied to train our models.
We train our models for 100K, 40K, 80K iterations on ADE20K, Cityscapes, PASCAL-Context datasets, respectively.

For the data augmentation, we randomly horizontally flip, randomly scale and crop the input images to a fixed size for training, which scales include $\{0.75, 1, 1.25, 1.5, 1.75, 2.0\}$.
Meanwhile, the cropped resolutions are $520 \times 520$ for ADE20K and PASCAL-Context, and $769 \times 769$ for Cityscapes dataset.

\paragraph{Inference.} In the inference phase, we adopt the sliding-window evaluation strategy~\cite{Zhao-CVPR-PSPNet-2017, Zhao-ECCV-PSANet-2018, Yu-CVPR-DFN-2018}.
Moreover, multi-scale and flipped inputs are employed to improve the performance, which scales contain $\{$0.5, 0.75, 1.0, 1.25$\}$ for the ADE20K and PASCAL-Context datasets, and $\{$0.5, 0.75, 1, 1.5$\}$ for the Cityscapes dataset.

\begin{table}[t]\centering
\caption{\textbf{Ablations on ADE20K}. We show mean IoU (\%) and pixel accuracy (\%) as the segmentation performance}
\label{tab:ablations}
\resizebox{\textwidth}{!}{
\subfloat[\textbf{Representative nodes}: 1 RepGraph layer of diverse nodes is inserted after the last stage of R50 baseline \label{tab:ablation:nodes}]{
\tablestyle{4pt}{1.05}
\begin{tabular}{r|x{22}x{22}}
\shline
\multicolumn{1}{c|}{model, R50}  & mIoU & pixAcc \\
\shline
R50 baseline & 36.48 & 77.57 \\
NL baseline & 40.97  & 79.96 \\
\hline
$S=1$ & 42.53 & 80.08   \\
$9$ & \bd{43.12} & 80.27  \\
$12$ & 42.60 & 80.43  \\
$15$ & 42.99 & 80.45  \\
$18$ & 42.85 & 80.44  \\
$27$ & 43.06 & \bd{80.64}  \\
\shline
\end{tabular}}\hspace{2mm}
\subfloat[\textbf{Instantiations}: 1 RepGraph layer of different structure is inserted after the last stage of ResNet-50 baseline \label{tab:ablation:instantiations}]{
\tablestyle{3pt}{1.05}
\begin{tabular}{ll|x{22}x{22}}
\shline
\multicolumn{2}{c|}{model, R50} & mIoU & pixAcc \\
\shline
baseline &  & 36.48 & 77.57 \\
\hline
\multirow{2}{*}{+NL} & sum & 40.3 & 79.96   \\
& concate & 40.97  & 80.01  \\
\hline
\multirow{2}{*}{+Simple} & sum & 42.61 & 80.191  \\
& concate  & 42.98 & 80.41 \\
\hline
\multirow{2}{*}{+Bottleneck} & sum  & 42.67  & \textbf{80.46} \\
& concate  & \textbf{43.12} & 80.27 \\
\shline
\multicolumn{3}{c}{} \\ %
\end{tabular}}\hspace{2mm}
\subfloat[\textbf{Stages}: 1 Bottleneck RepGraph layer is inserted into different stages of ResNet-50 baseline  \label{tab:ablation:stages}]{
\tablestyle{3pt}{1.05}
\begin{tabular}{c|x{22}x{22}}
\shline
\multicolumn{1}{c|}{model, R50}  & mIoU & pixAcc\\
\shline
baseline & 36.48 & 77.57 \\
\hline
res$_2$ & 41.52	& 79.67 \\
res$_3$ & \bd{41.59} & \bd{79.97}  \\
res$_4$ & 41.25 & 79.69 \\
res$_5$ & 41.34 & 79.89  \\
\shline
 \multicolumn{3}{c}{~}\\
 \multicolumn{3}{c}{~}\\
 \multicolumn{3}{c}{~}\\
\end{tabular}}\hspace{2mm}
\subfloat[\textbf{Deeper non-local models}: we insert 1, 3, and 5 Bottleneck RepGraph layer into the ResNet-50 baseline \label{tab:ablation:deeper}]{
\tablestyle{4pt}{1.05}
\begin{tabular}{l|x{22}x{22}}
\shline
model, R50 & mIoU & pixAcc \\
\shline
baseline  & 36.48  & 77.57 \\
1-layer   & 41.59  & \bd{79.97} \\
3-layer   & \bd{41.86}  & 79.85 \\
5-layer   & 41.76  & 79.89 \\
\shline
 \multicolumn{3}{c}{~}\\
 \multicolumn{3}{c}{~}\\
 \multicolumn{3}{c}{~}\\
  \multicolumn{3}{c}{~}\\
\end{tabular}}}
\end{table}

\subsection{Ablative Evaluation on ADE20K}
\label{sec:ablation}
This section provides an ablative evaluation on ADE20K comparing segmentation accuracy and computation complexity.
We train all models on the training set and evaluate on the validation set.
We adopt the pixel accuracy (pixAcc) and mean intersection of union (mIoU) as the evaluation metric.

\paragraph{Baselines.}
Similar to \cite{Zhao-CVPR-PSPNet-2017, Zhao-ECCV-PSANet-2018, Zhang-CVPR-EncNet-2018, Chen-Arxiv-Deeplabv2-2016, Chen-Arxiv-Deeplabv3-2017}, we adopt dilated ResNet (ResNet-50)~\cite{He-CVPR-ResNet-2016} with pre-trained weights as our \emph{backbone baseline}.
An auxiliary loss function with the weight of $0.4$ is integrated into the fourth stage of the backbone network~\cite{Yu-CVPR-CPN-2020, Zhao-CVPR-PSPNet-2017, Zhao-ECCV-PSANet-2018}.
We utilize one $3 \times 3$ convolution layer followed by batch normalization~\cite{Ioffe-ICML-BN-2015} and ReLU activation function on the output of the last backbone stage to reduce the channel dimension to 512.
Based on this output, we apply one non-local block (NL) as the \emph{NL baseline}.
Table~\ref{tab:ablations} (a) shows the segmentation performance of backbone baseline and NL baseline.

\paragraph{Instantiations.}
Table~\ref{tab:ablations} (b) shows the different structures of RepGraph layer, as illustrated in Figure~\ref{fig:structure}.
The simple RepGraph layer (SRG) has a similar structure with the non-local operation, while the bottleneck RepGraph layer (BRG) combines the residual bottleneck~\cite{He-CVPR-ResNet-2016, Xie-CVPR-ResNeXt-2017} with the simple RepGraph layer.
The number of sampled representative nodes is $9$.
Meanwhile, we also compare the different fusion methods (\emph{summation} and \emph{concatenation}) of diverse structures.
The bottleneck RepGraph layer has stronger representation ability, which achieves better performance than the simple RepGraph layer.
Therefore, in the rest of this paper, we use the bottleneck RepGraph layer version by default.

\begin{table}[t]
\centering
\small
\tablestyle{5pt}{1.03}
\caption{Practical GFLOPs of different blocks with the input feature of $256 \times 128$ resolution ($1/8$ of the $1024 \times 2048$ image). The batch size of the input feature is $1$, while the input channel $C = 2048$ and middle channel $C' = 256$. The inference time is measured on one NVIDIA RTX 2080Ti card. The decrease of our methods in term of computation and inference time is compared with NL}
\label{tab:complexity}
\begin{tabular}{l|l|l|l}  
\shline
\multicolumn{1}{c|}{input size} & \multicolumn{1}{c|}{model}  & \multicolumn{1}{c|}{GFLOPs} & \multicolumn{1}{c}{Inference Time(ms)}\\
\shline
 \multirow{4}{*}{$256 \times 128$} & NL~\cite{Yuan-Arxiv-OCNet-2018} & 601.4 & 146.65 \\
 & DANet~\cite{Fu-CVPR-DANet-2019} & 785.01 & 279.56 \\
 & SRG [ours] & 45.31 ($\downarrow 556.09$) & 60.89 ($\downarrow 85.76$)\\
 & BRG [ours] & 34.96 ($\downarrow 566.44$) & 25.96 ($\downarrow 120.69$)\\ 
 \shline
 \end{tabular}
\end{table}

\begin{table}[t]\centering
\caption{\textbf{Extended instantiations of RepGraph layer}. We can spatially group the spatial nodes, termed \emph{Grid RepGraph}, or divide the channel into several groups, termed \emph{Group RepGraph}. We show mean IoU (\%) and pixel accuracy (\%) as the segmentation performance}
\label{tab:group}
\subfloat[\textbf{Spatial group}: $g_s$ is the number of spatially grouped elements in one dimension, (\eg for a 2D input, $g_s=5$ indicates groups $5 \times 5$ elements as a graph node)\label{tab:group:spatial}]{
\tablestyle{6pt}{1.03}
\begin{tabular}{r|x{22}x{22}}
\shline
\multicolumn{1}{c|}{model, R50}  & mIoU & pixAcc \\
\shline
\multicolumn{1}{l|}{baseline} & 36.48 & 77.57\\
\multicolumn{1}{l|}{Grid RG($g_s=1$)} & 43.12 & 80.27\\
\hline
$g_s=5$ &  42.40 & 80.17   \\
$13$ &     41.82 & 80.01  \\
$65$ &     41.23 & 79.73  \\
\shline
\multicolumn{3}{c}{} \\ %
\end{tabular}}\hspace{5mm}
\subfloat[\textbf{Channel group}: $g_c$ indicates how many channels require dividing into on group
\label{tab:group:channel}]{
\tablestyle{5pt}{1.03}
\begin{tabular}{r|x{22}x{22}}
\shline
\multicolumn{1}{c|}{model, R50}  & mIoU & pixAcc \\
\shline
\multicolumn{1}{l|}{baseline} & 36.48 & 77.57\\
\multicolumn{1}{l|}{Group RG($g_c=1$)} & 43.12 & 80.27 \\
\hline
$g_c=4$ &  42.78 & 80.20   \\
$8$ &      43.01 & 80.32   \\
$16$ &     42.96 & 80.19   \\
$32$ & 	   42.38 & 80.18   \\
\shline
\end{tabular}}
\end{table}

\paragraph{How many representative nodes to sample?}
Table~\ref{tab:ablations} (a) compares the performance of choosing different number of representative nodes.
It shows \emph{all} the models can improve the performance over the ResNet-50 baseline and are better than the NL baseline, which validates the effectiveness and robustness of our RepGraph layer.
We employ $S=9$ as our default.
Interestingly, even only choosing one representative node $(s=1)$ for each position can also lead to a $1.56\%$ performance improvement.
This validates reducing redundancy in non-local helps to more effective representation.
For better understanding, we show some visualization of learned sampling positions in the supplementary material.

Next, we investigate the combination with the pre-trained model (\eg ResNet).
Due to insertion into the pre-trained model, we can not change the initial behaviour of the pre-trained model.
Therefore, we have to choose the summation version of bottleneck RepGraph layer and remove the last ReLU function of this layer.
Meanwhile, the parameters of the last convolution layer and batch normalization require to initialize as zero.

\paragraph{Which stage to insert RepGraph layer?} 
We insert the bottleneck RepGraph layer before the last block of different backbone stages, as shown in Table~\ref{tab:ablations} (c).
The improvements over the backbone baseline validate the RepGraph layer can be a generic component to extract features.

\paragraph{Going deeper with RepGraph layer.} 
Table~\ref{tab:ablations} (d) shows the performance with more RepGraph layers.
We add 1 (to res$_3$), 3 (to res$_3$, res$_4$, res$_5$ respectively), and 5 (1 to res$_3$, 2 to res$_4$, 2 to res$_5$).
With more layers, the performance can be improved further.
This improvement validates the RepGraph layer can model some complementary information not encoded in the pre-trained model. 

\paragraph{Computation complexity.}
The theoretical computational complexity of non-local operation and RepGraph layer is $\mathcal{O}(C \times N^2)$ and $\mathcal{O}(C \times S \times N)$ respectively. 
Meanwhile, Table~\ref{tab:complexity} shows the practical GFLOPs and inference time of non-local operation~\cite{Wang-CVPR-Nonlocal-2018}, DANet~\cite{Fu-CVPR-DANet-2019} and RepGraph layer with the input size of $256\times128$ ($1/8$ of the $1024 \times 2048$ image).
Here, we use the concatenation fusion method in each block.
The RepGraph layer can dramatically reduce the computation complexity and have fewer inference time compared to the non-local operation.

Then, we show some extended instantiations of RepGraph layer.

\begin{table}[t]
\centering
\small
\tablestyle{6pt}{1.03}
\caption{Quantitative evaluations on the ADE20K validation set. 
The proposed RGNet performs favorably against the \emph{state-of-the-art} segmentation algorithms}
\label{tab:perf:ade}
\begin{tabular}{l|c|l|x{22}|x{22}}
\shline
\multicolumn{1}{c|}{model} & reference & \multicolumn{1}{c|}{backbone} & \emph{mIoU} & \emph{picAcc} \\
\shline
RefineNet~\cite{Lin-CVPR-Refinenet-2017} & CVPR2017 & ResNet-152 & 40.7  &  -     \\
UperNet~\cite{Xiao-UperNet-ECCV-2018} 	 & ECCV2018 & ResNet-101 & 42.66 &  81.01 \\
PSPNet~\cite{Zhao-CVPR-PSPNet-2017}      & CVPR2017 & ResNet-269 & 44.94 &  81.69 \\
DSSPN~\cite{Liang-DSSPN-CVPR-2018}       & CVPR2018 & ResNet-101 & 43.68 &  81.13 \\
PSANet~\cite{Zhao-ECCV-PSANet-2018}      & ECCV2018 & ResNet-101 & 43.77 &  81.51 \\
SAC~\cite{Zhang-ICCV-SAC-2017} 			 & ICCV2017 & ResNet-101 & 44.30 &  \textbf{81.86} \\
EncNet~\cite{Zhang-CVPR-EncNet-2018}     & CVPR2018 & ResNet-101 & 44.65 &  81.69 \\
CFNet~\cite{Zhang-CVPR-ACF-2019} 		 & CVPR2019	& ResNet-101 & 44.89 & - \\
CCNet~\cite{Huang-ICCV-CCNet-2019}	 	 & ICCV2019 &
	ResNet-101 & 45.22 & - \\
ANL~\cite{Zhu-ICCV-ANL-2019}  			 & ICCV2019	& ResNet-101 & 45.24 & - \\
DMNet~\cite{He-ICCV-DMNet-2019}			 & ICCV2019 &
	ResNet-101 & \underline{45.50} \\
\hline
RGNet  & - & ResNet-50  & 44.02 & 81.12\\
RGNet & - & ResNet-101  & \bd{45.8} & \underline{81.76}  \\
\shline
\end{tabular}
\end{table}

\begin{table}[t]
\centering
\small
\tablestyle{5pt}{1.03}
\caption{
Quantitative evaluations on Cityscapes test set. 
The proposed RGNet performs favorably against the \emph{state-of-the-art} segmentation methods.
We train our model with \emph{trainval-fine} set, and evaluate on the \emph{test} set
}
\label{tab:perf:cityscapes}
\begin{tabular}{l|c|l|x{36}}
\shline
\multicolumn{1}{c|}{model}  & reference & \multicolumn{1}{c|}{backbone} & \emph{mIoU} \\
\shline
GCN~\cite{Peng-CVPR-Largekernl-2017} 	  & CVPR2017 & ResNet-101 & 76.9 \\
DUC~\cite{Wang-WACV-DUC-2018} 		      & WACV2018 & ResNet-101 & 77.6 \\
DSSPN~\cite{Liang-DSSPN-CVPR-2018} 		  & CVPR2018 & ResNet-101 & 77.8 \\
SAC~\cite{Zhang-ICCV-SAC-2017} 		      & ICCV2017 & ResNet-101 & 78.1 \\
PSPNet~\cite{Zhao-CVPR-PSPNet-2017} 	  & CVPR2017 & ResNet-101 & 78.4 \\
BiSeNet~\cite{Yu-ECCV-BiSeNet-2018} 	  & ECCV2018 & ResNet-101 & 78.9 \\
AAF~\cite{Ke-ECCV-AAF-2018} 			  & ECCV2018 & ResNet-101 & 79.1 \\
DFN~\cite{Yu-CVPR-DFN-2018} 			  & CVPR2018 & ResNet-101 & 79.3 \\
PSANet~\cite{Zhao-ECCV-PSANet-2018} 	  & ECCV2018 & ResNet-101 & 80.1 \\
DenseASPP~\cite{Yang-CVPR-DenseASPP-2018} & CVPR2018 & DenseNet-161 & 80.6 \\
ANL~\cite{Zhu-ICCV-ANL-2019} 			  & ICCV2019 & ResNet-101 & 81.3 \\
CPNet~\cite{Yu-CVPR-CPN-2020}				  & CVPR2020 & ResNet-101 & 81.3 \\
CCNet~\cite{Huang-ICCV-CCNet-2019}	 	 & ICCV2019 &
	ResNet-101 & \underline{81.4} \\
DANet~\cite{Fu-CVPR-DANet-2019}           & CVPR2019 & ResNet-101 & \bd{81.5} \\
\hline
RGNet & - & ResNet-101 & \bd{81.5}   \\
\shline
\end{tabular}
\end{table}

\paragraph{Extension.}
Table~\ref{tab:group} shows some extended instantiations of the RepGraph layer.
Instead of considering one feature element as a graph node, we can spatially group a few pixels as a graph node to construct the RepGraph layer, termed \emph{Grid RepGraph}.
We argue that the Grid RepGraph layer computes the relationships between representative nodes and local nodes in one group.
The sampling of representative nodes can capture long-range information, while the spatial grouping enables the short-range contextual modelling.

Inspired by \cite{Yue-NIPS-CGNL-2018, Xie-CVPR-ResNeXt-2017}, the channel of RepGraph layer can be divided into a few groups, called \emph{Group RepGraph} layer.
This structure can increase the cardinality and capture the correlation in diverse channel groups.
Although there is a little performance decrease, the extended instantiations are more efficient.

\begin{table}[t]
\centering
\scriptsize
\tablestyle{5pt}{1.03}
\caption{
Quantitative evaluations on the PASCAL-Context validation set. 
The proposed RGNet performs favorably against the \emph{state-of-the-art} segmentation methods.
}
\label{tab:perf:pascal-context}
\begin{tabular}{l|c|l|x{36}}
\shline
\multicolumn{1}{c|}{model}  & reference & \multicolumn{1}{c|}{backbone} & \emph{mIoU} \\
\shline
CRF-RNN~\cite{Zheng-ICCV-CRFasRNN-2015} 			   & ICCV2015 & VGG-16 & 39.3 \\
RefineNet~\cite{Lin-CVPR-Refinenet-2017}         	   & CVPR2017 & ResNet-152 & 47.3 \\
PSPNet~\cite{Zhao-CVPR-PSPNet-2017} 		           & CVPR2017 & ResNet-101 & 47.8 \\
CCL~\cite{Ding-CVPR-CCL-2018} 					       & CVPR2018 & ResNet-101 & 51.6 \\
EncNet~\cite{Zhang-CVPR-EncNet-2018} 				   & CVPR2018 & ResNet-101 & 51.7 \\
DANet~\cite{Fu-CVPR-DANet-2019} 					   & CVPR2019 & ResNet-101 & 52.6 \\
ANL~\cite{Zhu-ICCV-ANL-2019} 						   & ICCV2019 & ResNet-101 & 52.8 \\
EMANet~\cite{Li-ICCV-EMANet-2019}					   &
	ICCV2019 & ResNet-101 & \underline{53.1} \\
CPNet~\cite{Yu-CVPR-CPN-2020}							   & CVPR2020 & ResNet-101 & \bd{53.9} \\
\hline
RGNet & - & ResNet-101 	   &  \bd{53.9}  \\
\shline
\end{tabular}
\end{table}

\begin{table}[t]
\centering
\scriptsize
\caption{Adding 1 RepGraph layer to Mask R-CNN for COCO \bd{object detection} and \bd{instance segmentation}. The backbone is ResNet-50 with FPN \cite{Lin-CVPR-FPN-2017}}
\label{tab:coco_det}
\tablestyle{4pt}{1.03}
\begin{tabular}{cc|x{22}x{22}x{22}|x{22}x{22}x{22}}
\shline
 \multicolumn{2}{c|}{method} & AP$^\text{box}$ & AP$^\text{box}_{50}$ & AP$^\text{box}_{75}$
 & AP$^\text{mask}$ & AP$^\text{mask}_{50}$ & AP$^\text{mask}_{75}$ \\[.1em]
\shline
\multirow{3}{*}{R50} & baseline & 38.0 & 59.6 & 41.0 & 34.6 & 56.4 & 36.5 \\
& +1 NL & 39.0 & 61.1 & 41.9 & 35.5 & \bd{58.0} & 37.4\\
& +1 RGL & \bd{39.6} & \bd{61.4}  & \bd{42.1}  & \bd{36.0}  & 57.9  & \bd{37.9}  \\
\shline
\end{tabular}
\end{table}

\subsection{Performance Evaluation}
\label{sec:performance}
In this section, we compare the RepGraph network (RGNet) with other \emph{state-of-the-art} methods on three datasets: ADE20K, Cityscapes, and PASCAL-Context.

\paragraph{ADE20K.}
Table~\ref{tab:perf:ade} shows the comparison results with other \emph{state-of-the-art} algorithms on ADE20K dataset.
\emph{Without any bells and whistles}, our RGNet with ResNet-101 as backbone achieves mean IoU of $45.8\%$ and pixel accuracy of $81.76\%$, which outperforms previous \emph{state-of-the-art} methods.
Our RGNet with ResNet-50 obtains mean IoU of $44.04\%$ and pixel accuracy of $81.12\%$, even better than the PSANet~\cite{Zhao-ECCV-PSANet-2018}, PSPNet~\cite{Zhao-CVPR-PSPNet-2017}, UperNet~\cite{Xiao-UperNet-ECCV-2018}, and RefineNet~\cite{Lin-CVPR-Refinenet-2017} with deeper backbone networks.

\paragraph{Cityscapes.}
Table~\ref{tab:perf:cityscapes} shows the comparison with previous results on Cityscapes~\cite{Cityscapes} dataset.
We train our model with \emph{trainval} set of merely the fine annotation images, and evaluate on the \emph{test} set.
The compared methods only use the fine-annotation images as well.
The RGNet achieves mean IoU of $81.5\%$, which competes with previous \emph{state-of-the-art} methods.
However, as shown in Table~\ref{tab:perf:cityscapes}, the RGNet is more efficient than the DANet, which applies the self-attention mechanism on the spatial and channel dimension respectively.

\paragraph{PASCAL-Context.}
Table~\ref{tab:perf:pascal-context} shows the results on the PASCAL-Context dataset compared with other methods.
The RGNet achieves mean IoU of $53.9\%$ on the \emph{val} set, which sets \emph{state-of-the-art} result.

\section{Experiments on Detection}
\label{sec:exp-ext}
To investigate the generalization ability of our work, we conduct experiments on object detection.
Following \cite{Wang-CVPR-Nonlocal-2018}, we set the Mask R-CNN~\cite{He-ICCV-MaskRCNN-2017} as our baseline.
All experiments are trained on COCO~\cite{Lin-COCO-2014} \emph{train}2017 and tested on \emph{test}2017.

We add one RepGraph layer before the last block of res$_4$ of ResNet backbone network in the Mack R-CNN. 
Table~\ref{tab:coco_det} shows the box AP and mask AP on COCO dataset.
As we can see, using just one RepGraph layer can improve the performance over the baseline.
Meanwhile, adding one RepGraph layer achieves \emph{better} performance than adding one non-local operation.

\section{Concluding Remarks}
\label{sec:conclusion}
We present a Representative Graph (RepGraph) layer to model long-range dependencies via dynamically sample a few representative nodes.
The RepGraph layer is compact and general component for visual understanding.
Meanwhile, the RepGraph layer is easy to integrate into any pre-trained model or combined with other designs.
On the semantic segmentation and object detection task, the RepGraph layer can achieve promising improvement over baseline and non-local operation.
We believe the \textit{RepGraph} layer can be an efficient and general block to the visual understanding community.

\paragraph*{Acknowledgment:}
This work is supported by the National Natural Science Foundation of China (No.61433007 and 61876210).

\section*{Appendix}
\subsection*{Visualization of representative nodes}
\begin{figure}[h]
\scriptsize
\hsize=\textwidth
\centering
\tablestyle{1pt}{0.5}
\begin{center}
\begin{tabular}{ccccc}
\includegraphics[width=0.2\textwidth]{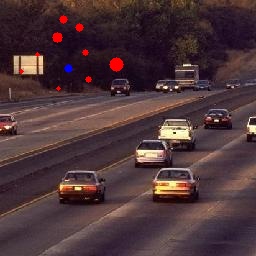}  &
\includegraphics[width=0.2\textwidth]{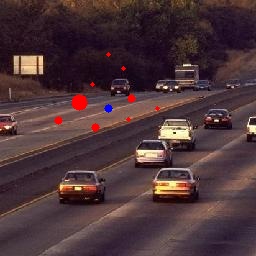}  &
\includegraphics[width=0.2\textwidth]{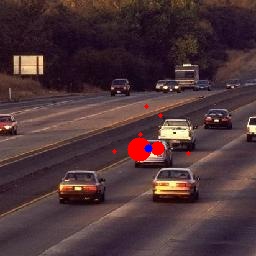}  &
\includegraphics[width=0.2\textwidth]{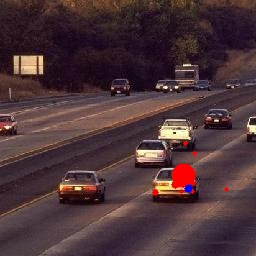}  &
\includegraphics[width=0.2\textwidth]{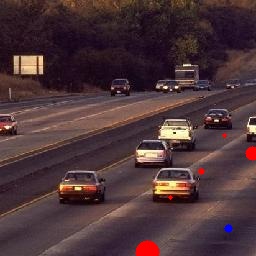}  \\
\includegraphics[width=0.2\textwidth]{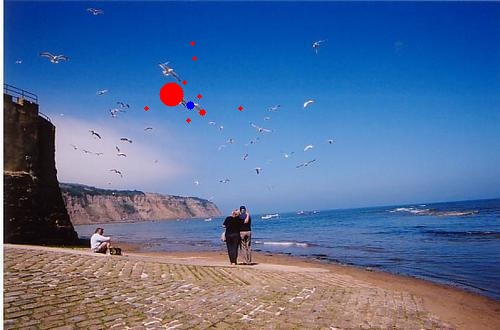}  &
\includegraphics[width=0.2\textwidth]{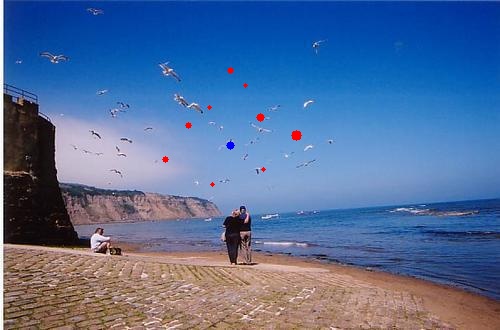}  &
\includegraphics[width=0.2\textwidth]{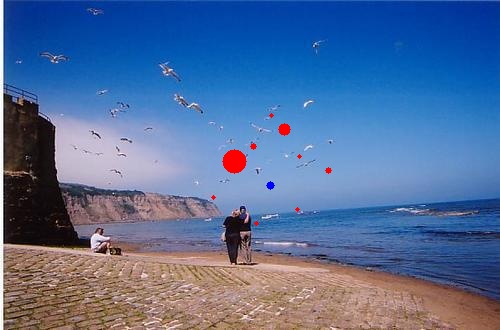}  &
\includegraphics[width=0.2\textwidth]{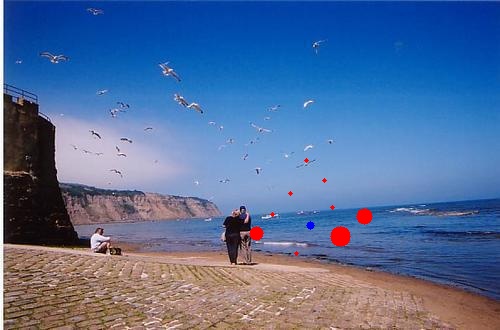}  &
\includegraphics[width=0.2\textwidth]{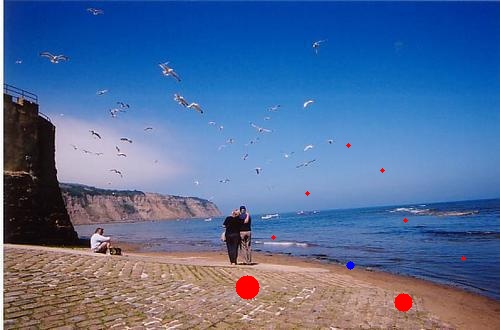}  \\
\includegraphics[width=0.2\textwidth]{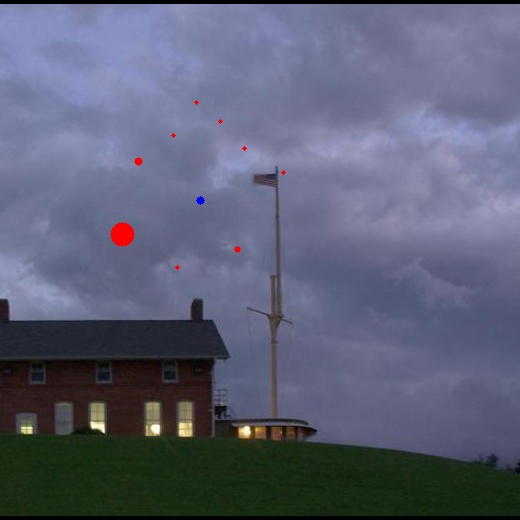}  &
\includegraphics[width=0.2\textwidth]{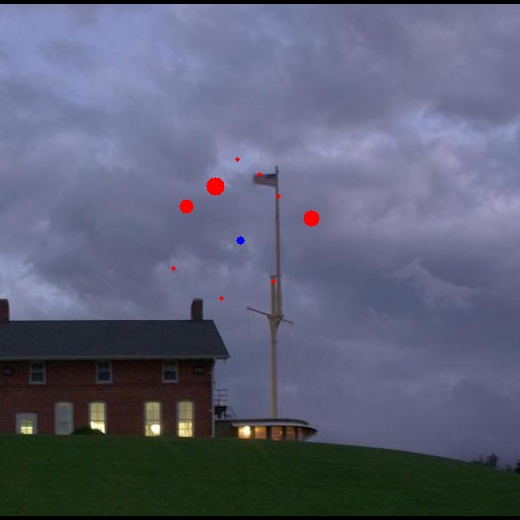}  &
\includegraphics[width=0.2\textwidth]{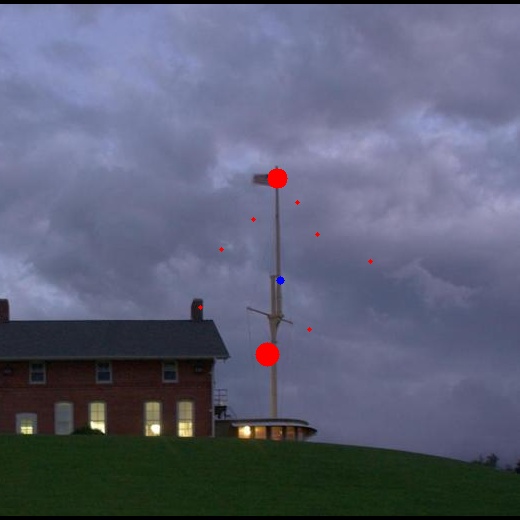}  &
\includegraphics[width=0.2\textwidth]{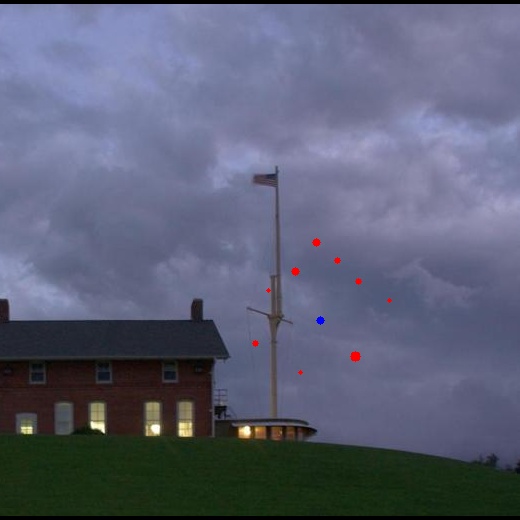}  &
\includegraphics[width=0.2\textwidth]{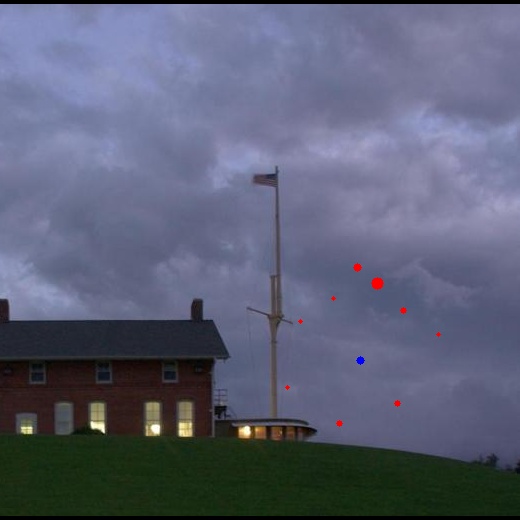}  \\
\includegraphics[width=0.2\textwidth]{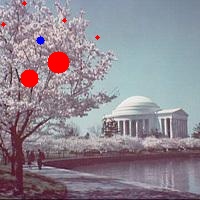}  &
\includegraphics[width=0.2\textwidth]{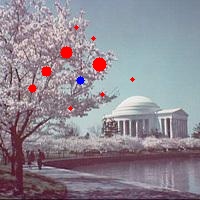}  &
\includegraphics[width=0.2\textwidth]{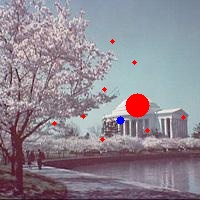}  &
\includegraphics[width=0.2\textwidth]{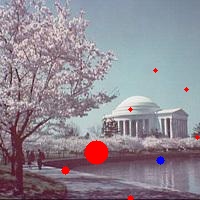}  &
\includegraphics[width=0.2\textwidth]{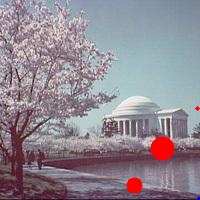}  \\
\includegraphics[width=0.2\textwidth]{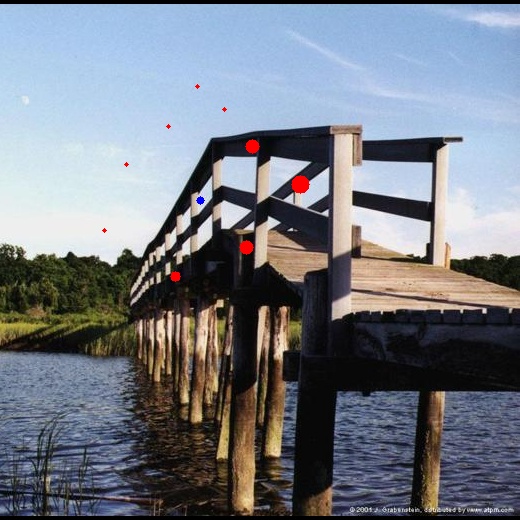}  &
\includegraphics[width=0.2\textwidth]{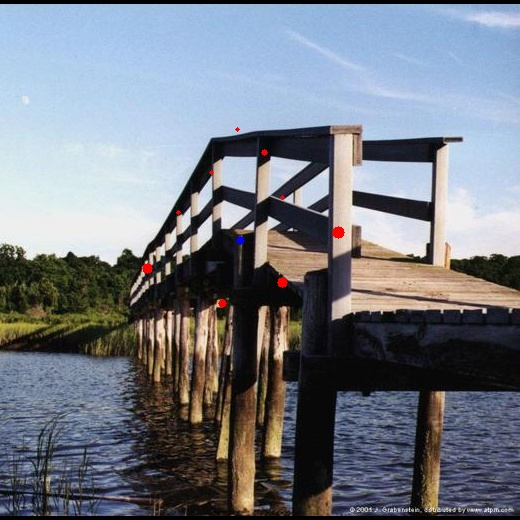}  &
\includegraphics[width=0.2\textwidth]{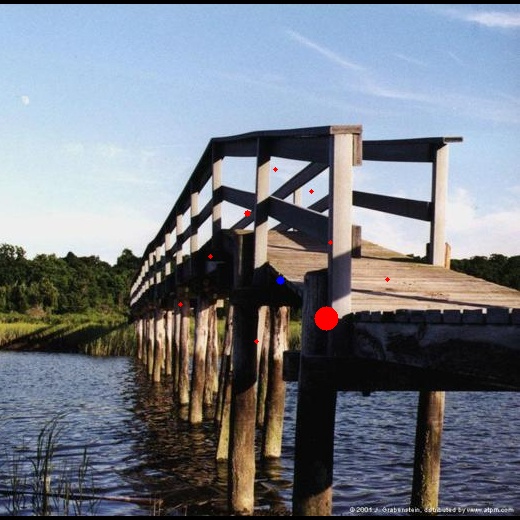}  &
\includegraphics[width=0.2\textwidth]{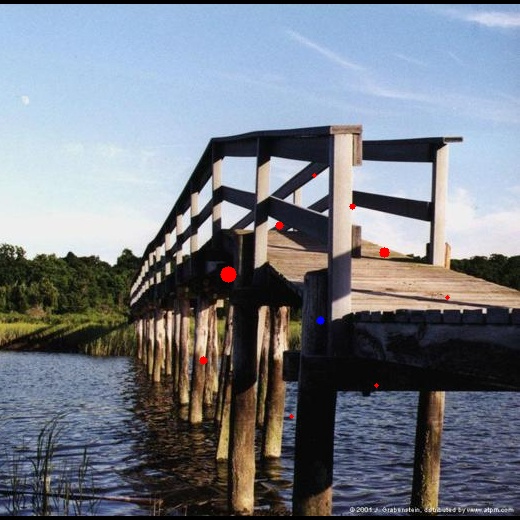}  &
\includegraphics[width=0.2\textwidth]{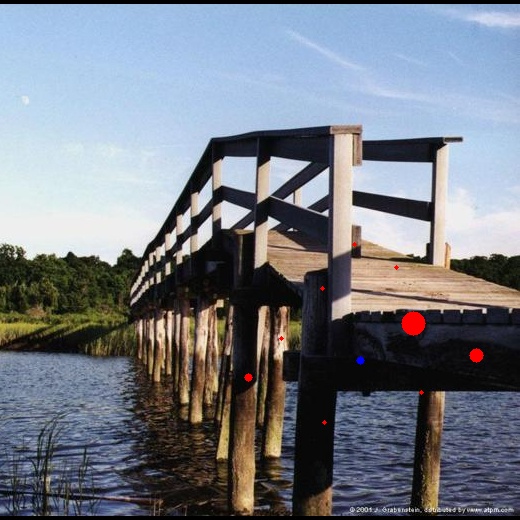}  \\
\includegraphics[width=0.2\textwidth]{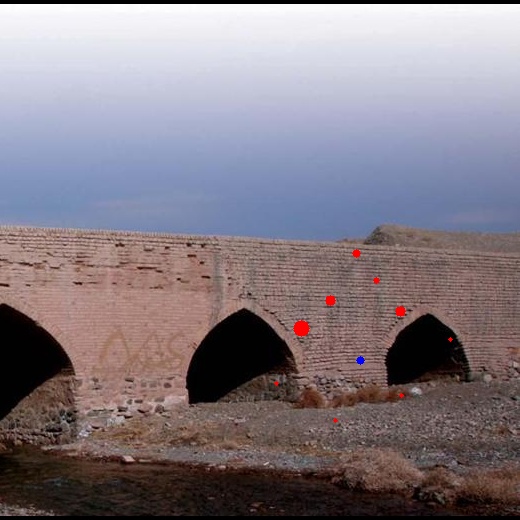}  &
\includegraphics[width=0.2\textwidth]{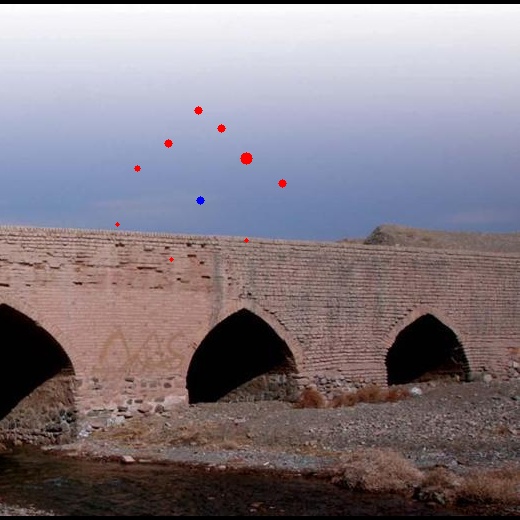}  &
\includegraphics[width=0.2\textwidth]{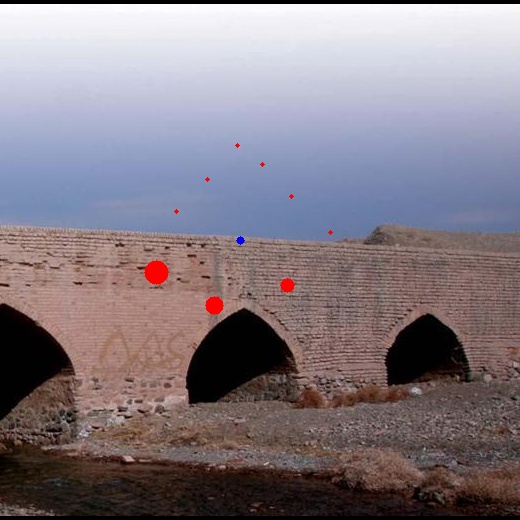}  &
\includegraphics[width=0.2\textwidth]{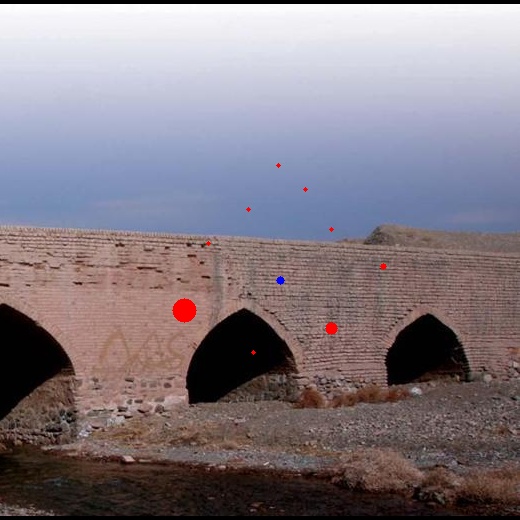}  &
\includegraphics[width=0.2\textwidth]{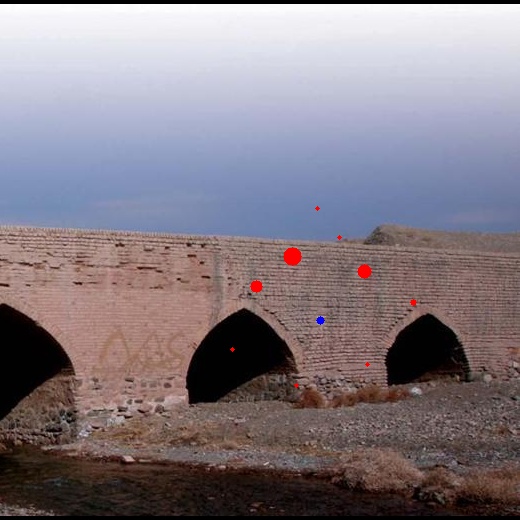}  \\
\end{tabular}
\end{center}
\caption{\textbf{Visualization of representative nodes for different positions.} The {\color{blue}{blue}} point indicates the current position, while the {\color{red}{red}} points denotes the sampled positions. The sizes of the {\color{red}{red}} points represents the weight of this sampled point. For better view, we adjust the size of points which has very small weights.}
\label{fig:visualization}
\end{figure}

The proposed Representative Graph (RepGraph) layer can sample a few representative nodes to dramatically reduce the computational complexity of non-local operation~\cite{Wang-CVPR-Nonlocal-2018, Hu-CVPR-RelationNet-2018, Yuan-Arxiv-OCNet-2018}.
This layer applies the learned offset matrix to the original feature to dynamically sample the representative positions for each node.
The representative features enable a compact and effective representation, which improves the performance in semantic segmentation and object detection.
Here, for better understanding, we visualize the learned representative positions in Figure~\ref{fig:visualization}. 
The RepGraph layer is applied on the ADE20K dataset of the semantic segmentation task. 

As shown in Figure~\ref{fig:visualization}, the proposed RepGraph layer can capture representative features for different positions.
In the first row, the representative positions can focus on the \emph{car}, while the \emph{current} position is on the \emph{car}.
When the \emph{current} position is on the \emph{road}, the receptive field of the representative positions becomes larger to capture the \emph{large} stuff.
In the second row, while the \emph{current} position is on a \emph{small} \emph{Seagull}, the RepGraph layer can also capture the representative features.
For the \emph{slender} \emph{rod} in the third row, the RepGraph layer can model the representative positions on it.
Meanwhile, for the \emph{hollow} things, \eg the \emph{bridge} in the fifth row, the learned representative positions can be on the hollow bridge.
When on the boundary of some things, \eg the boundary of the \emph{bridge} in the last row, the RepGraph layer can model the representative features of the \emph{bridge} to recognize it.

These visualizations can improve the understanding of the RepGraph layer and validate the effectiveness of the RepGraph layer.
We believe the proposed RepGraph layer can be helpful to the visual understanding community.

\clearpage
\bibliographystyle{splncs04}
\bibliography{reference}
\end{document}